%% file: main.tex
\documentclass[10pt, conference]{IEEEtran}

\ifCLASSINFOpdf
   \usepackage[pdftex]{graphicx}
\else
\fi
\usepackage{lettrine}
\usepackage{wrapfig}
\usepackage{lipsum}
\usepackage{caption}
\usepackage{subcaption}
\usepackage{booktabs}
\usepackage{hhline}
\usepackage{algorithm,algorithmic}
\usepackage{amsmath,amsfonts}
\usepackage{hyperref}
\usepackage{varioref}
\hypersetup{draft}
\usepackage{comment}
\usepackage{float}
\usepackage{upgreek} 
\usepackage{multirow,array}

\usepackage{longtable}

\def\BibTeX{{\rm B\kern-.05em{\sc i\kern-.025em b}\kern-.08em
    T\kern-.1667em\lower.7ex\hbox{E}\kern-.125emX}}
\providecommand{\keywords}[1]
{
  \small	
  \textbf{\textit{Keywords---}}\textbf{\textit{#1}}
}

\makeatletter

\title{Comparison of Various SLAM Systems for Mobile Robot in an Indoor Environment}

\begin{document}

\author{
\IEEEauthorblockN{Maksim~Filipenko}
\IEEEauthorblockA{\textit{Institute of Robotics} \\ \textit{Innopolis University} \\ Innopolis, Russia \\
m.filipenko@innopolis.ru}
\and
\IEEEauthorblockN{Ilya Afanasyev}
\IEEEauthorblockA{\textit{Institute of Robotics} \\ \textit{Innopolis University}\\ Innopolis, Russia \\
i.afanasyev@innopolis.ru}}
\maketitle

\input{Sections/abstract}
\input{Sections/introduction}
\input{Sections/related}

\input{Sections/system}

\input{Sections/experiment}
\input{Sections/evaluation}
\input{Sections/conclusion}

\section*{Acknowledgements}
This research has been supported by the Russian Ministry of Education and Science within the Federal Target Program grant (research grant ID RFMEFI60617X0007).



\bibliography{main}
\bibliographystyle{IEEEtran}

\end{document}

%% file: Sections/abstract.tex
\begin{abstract}
 This article presents a comparative analysis of a mobile robot trajectories computed by various ROS-based SLAM systems. For this reason we developed a prototype of a mobile robot with common sensors: 2D lidar, a monocular and ZED stereo cameras. Then we conducted  experiments in a typical office environment and collected data from all sensors, running all tested SLAM systems based on the acquired dataset. We studied the following SLAM systems: (a) 2D lidar-based: GMapping, Hector SLAM, Cartographer; (b) monocular camera-based: Large Scale Direct monocular SLAM (LSD SLAM), ORB SLAM, Direct Sparse Odometry (DSO); and (c) stereo camera-based: ZEDfu, Real-Time Appearance-Based Mapping (RTAB map), ORB SLAM, Stereo Parallel Tracking and Mapping (S-PTAM). Since all SLAM methods were tested on the same dataset we compared results for different SLAM systems with appropriate metrics, demonstrating encouraging results for lidar-based Cartographer SLAM, Monocular ORB SLAM and Stereo RTAB Map methods. \\
\end{abstract}

\keywords{simultaneous localization and mapping (SLAM), visual odometry, indoor navigation, ROS}

%% file: Sections/introduction.tex
\section{Introduction}
\label{introduction}

The map building problem for an unknown environment with use of onboard sensors while solving the localization problem at the same time is known as Simultaneous Localization and Mapping (SLAM) \cite{Leonard1991,cadena2016}. One of the classical approaches to solve this problem is filter-based SLAM. This method is based on the Bayesian filtering theory. It has two main steps: (1) the prediction step, when robot localization and map state are updated using the previous information about the system state and input control commands; (2) the measurement update, where current sensor data are matched to predicted system state in order to make new system state prediction. This approach has various implementations. The earlier SLAM system were based on Extended Kalman filter (EKF-based SLAM, \cite{bailey2006b}), or on Particle Filter (e.g., FastSLAM, \cite{montemerlo2002}). For introduction to the basics of SLAM system we direct to surveys \cite{Bailey2006}, for more detailed information about filter-based systems recommend \cite{Aulinas2008}.  

The SLAM problem can be solved with the use of different sensors and the choice of suitable sensors plays a special role in the effective operation of Unmanned Ground Vehicle (UGV) due to the limited autonomous resource. Most mobile robots have Inertial Measurement Unit (IMU), which includes accelerometers, gyroscope, and magnetometer, which measure orientation, angular velocity, and acceleration correspondingly. This approach to getting mobile robot localization at using only Inertial Navigation System (INS) can give significant navigation errors \cite{Barshan1995}. 
Therefore the main sensor for indoor robot navigation and SLAM is usually lidar. 2D lidar SLAM systems are currently presented in different packages like GMapping \cite{Grisetti2007} (which uses Rao-Blackwellized particle filer \cite{Murphy2000} to learn grid maps from 2D lidar data),  Hector SLAM \cite{Kohlbrecher2011} (that is another very popular ROS-based SLAM), and Cartographer \cite{hess2016} (which is one of the most recent systems).

On the other hand, monocular and stereo cameras are good low-cost passive sensors, which can effectively solve SLAM problem working as a single source of information about an environment that referring to Visual SLAM (V-SLAM \cite{Fuentes-Pacheco2012,Taketomi2017}). All methods of V-SLAM can be divided into two groups: (1) feature-based, which use selected features for map building, and (2) direct that work with entire images as a whole. The earlier investigations on visual navigation were carried out with a binocular stereo camera \cite{Se2002,Olson2003} and a monocular camera (that was called MonoSLAM \cite{Davison2003}).  Over the past decade, numerous methods have been proposed, including Parallel Tracking and Mapping (PTAM, \cite{klein2007,Bailey2006a}), REgularized MOnocular Depth Estimation (REMODE, \cite{Pizzoli2014}), Oriented FAST \cite{Rosten2006} and Rotated BRIEF \cite{Calonder2010}  (ORB-SLAM, \cite{Mur-Artal2015,Rublee2011}), Dense Tracking and Mapping in Real-Time (DTAM, \cite{Newcombe2011}), Large-Scale Direct monocular SLAM (LSD-SLAM, \cite{engel2014a}), Large-Scale Direct SLAM with Stereo Cameras (Stereo LSD-SLAM, \cite{engel2015}), Fast semi-direct monocular visual odometry (SVO,\cite{Forster2014}), Real-Time Appearance-Based Mapping (RTAB map, \cite{Labbe2014}), Dense Piecewise Parallel Tracking and Mapping (DPPTAM, \cite{Concha2015}), Direct Sparse Odometry (DSO, \cite{engel2018}), ElasticFusion (Dense SLAM without a Pose Graph \cite{Whelan2015}), Convolutional Neural Networks SLAM (CNN-SLAM, \cite{Tateno2017}), Stereo Parallel Tracking and Mapping (S-PTAM, \cite{Pire2017}). 
~
Each of these methods has its advantages and disadvantages. Common features include weak resistance to unfavorable conditions, experimentation, poor performance with a weak geometric diversity of the environment, sensitivity to pure rotations \cite{gauglitz2012}, \cite{Pirchheim2013}. These methods do not provide metric information that is required in some applications (this problem is also known as Scale ambiguity). The solution of this problem is discussed in computer vision and robotics society. Thus, e.g., the paper \cite{Nutzi2011} suggested to use additional sensors like IMU, whereas the authors  \cite{zhou2016} proposed the approach for restoring metric information using information about the geometric sensor arrangement. The other important things, which can contribute to SLAM quality, are map initialization \cite{Mulloni2013},\cite{Arth2015}, a loop closure \cite{newman2005} (that compensates errors accumulated during laps running by UGV) and a type of camera shutter, which is used \cite{Lovegrove2013}. 
More detailed description of features for Visual SLAM system \cite{younes2017}, which we investigated in this research,  is presented in the Table \ref{tab:slaminfo}.

\makeatletter
\let\@currsize\normalsize
\makeatother

\begin{table}[!htbp]
\captionsetup{font=scriptsize}
\caption{\label{tab:slaminfo}\fontsize{8}{12}\fontfamily{ptm}\selectfont \textsc{DESCRIPTION OF FEATURES FOR VISUAL SLAM SYSTEMS INVESTIGATED IN THIS RESEARCH}}
\centering
\begin{tabular}{ p{1.02cm} | p{0.7cm} | p{1cm} | p{1.25cm} | p{0.85cm} |p{1.4cm} }
\toprule
    System & Method & Feature type & Feature descriptor & Map density & Loop Closure \\
\midrule
        PTAM & feature & FAST & local patch of pixels & sparse & none \\  \hline
    SVO & semi-direct & FAST & local patch of pixels & sparse & none \\  \hline
    DPPTAM & direct & intensity gradient & local patch of pixels & dense & none \\ \hline
    LSD SLAM & direct & intensity gradient & local patch of pixels & dense & Bag of Words place recognition\\  \hline
    ORB SLAM & feature & FAST & ORB & semi-dense & FabMap\\  \hline
    DSO & direct & intensity gradient & local patch of pixels & dense & none \\  \hline
    RTAB map & feature & GFTT, FAST, ... & SIFT, ORB, ... & sparse & Bag of Words place recognition \\ \hline
    S-PTAM & feature & GFTT, FAST, ... & SIFT, ORB, ... & sparse & Bag of Words place recognition \\
\bottomrule
\end{tabular}
\end{table}

In this paper, we mainly focus on a real robotics application for existing SLAM systems, which are implemented in Robot Operating System (ROS, \cite{Quigley2009}). For this reason, we investigate the following SLAM systems: (a) 2D lidar-based: GMapping, Hector SLAM, Cartographer; (b) monocular camera-based: LSD SLAM, ORB SLAM, DSO; and (c) stereo camera-based: ZEDfu, RTAB map, ORB SLAM, S-PTAM.
The brief information about ROS-based SLAM methods which were studied in this research are presented in the Table \ref{tab:slamye}.
  
\begin{table}[!htbp]
    \captionsetup{font=scriptsize}
    \caption{\label{tab:slamye}\fontsize{8}{12}\fontfamily{ptm}\selectfont \textsc{ROS-BASED SIMULTANEOUS LOCALIZATION AND MAPPING (SLAM) SYSTEMS STUDIED IN THIS RESEARCH}}
    \centering
    \begin{tabular}{  p{0.6cm} | p{4.3cm} | p{1.5cm} | p{0.7cm} }
    \toprule
    \textbf {Year} & \textbf {System} & \textbf {Sensor} & \textbf {Ref.} \\
    \midrule
    2007 & GMapping & 2D lidar & \cite{Grisetti2007}\\ 
    2007 & Parallel Tracking and Mapping (PTAM) & mono & \cite{klein2007}\\ 
    2011 & Hector SLAM & 2D lidar & \cite{Kohlbrecher2011}\\ 
    2014 & Semi-direct Visual Odometry (SVO) & mono & \cite{Forster2014} \\
    2014 & Large Scale Direct monocular SLAM (LSD SLAM) & mono & \cite{engel2014a} \\
2014 & Real-Time Appearance-Based Mapping (RTAB map) & stereo & \cite{Labbe2014} \\
2015 & ORB SLAM & mono, stereo & \cite{Mur-Artal2015}\\
2015 & Dense Piecewise Parallel Tracking and Mapping (DPPTAM) & mono & \cite{Concha2015}\\
2016 & Direct Sparse Odometry (DSO) & mono & \cite{engel2018}\\
2016 & Cartographer & 2D lidar & \cite{hess2016} \\
2017 & Stereo Parallel Tracking and Mapping (S-PTAM) & stereo & \cite{Pire2017}\\
    \bottomrule
    \end{tabular}
\end{table}

  The paper is organized as follows: Section \ref{system} describes the hardware and software used, Section \ref{experiment} presents the methodology of the experiment and metrics for evaluation, Section \ref{evaluation} demonstrates the achieved results. Finally, we conclude in Section \ref{conclusion}.

%% file: Sections/related.tex
\section{Related work}
\label{related}

In this section we review the research papers dealt with comparative analysis of various ROS-based SLAM methods. 
There are plenty of recent investigations, which compare SLAM methods for indoor navigation in mobile robotics applications \cite{huletski2015,buyval2017,ibragimov2017,sokolov2017,giubilato2018}. Some studies compare only efficiency of lidar-based SLAM approaches \cite{kummerle2009,santos2013,rojas2018,yagfarov2018}. The paper \cite{santos2013} compares five ROS-based Laser-related SLAM methods: Gmapping, KartoSLAM, Hector SLAM, CoreSLAM and LagoSLAM, demonstrating more favourable outcomes for the last three.

The paper \cite{buyval2017} describes the qualitative comparison of monocular SLAM methods (ORB-SLAM, REMODE, LSD-SLAM, and DPPTAM) performed for indoor navigation. The research \cite{ibragimov2017} extends the comparative analysis, considering a mobile platform trajectories calculated from different sensors data (mono and stereo cameras, lidar, and Kinect 2 depth sensor) with various ROS-based SLAM methods in homogeneous indoor environment. The author used monocular ORB-SLAM, DPPTAM, stereo ZedFu and Kinect-related RTAB map packages, verifying the trajectories with a ground truth dealt with lidar-based Hector SLAM.

%% file: Sections/system.tex
\section{System setup}
\label{system}

\begin{figure}[!htbp]
    \captionsetup{font=scriptsize}
    \centering
    \includegraphics[width=0.7\linewidth]{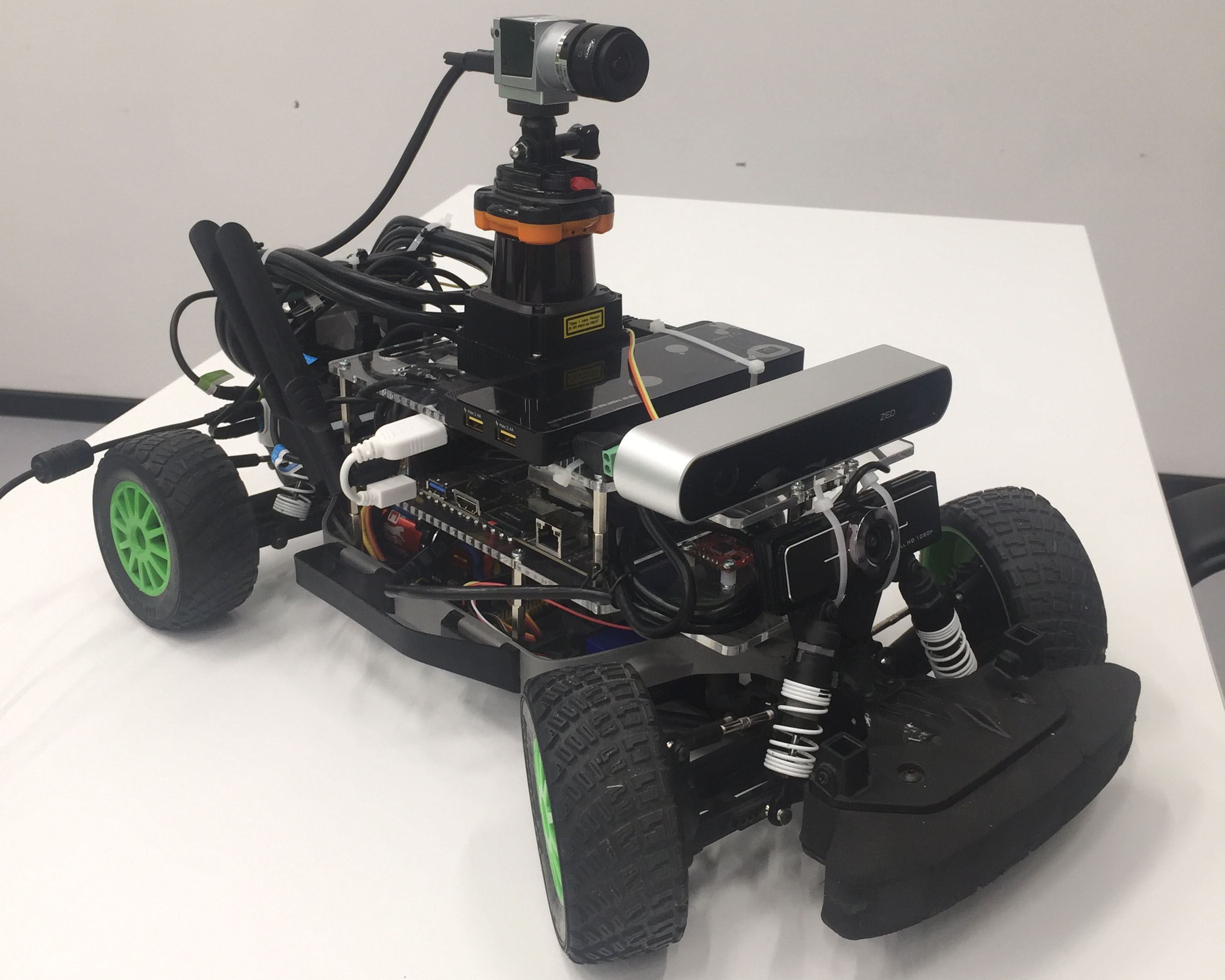}
    \caption{\fontsize{8}{12}\fontfamily{ptm}\selectfont Innopolis UGV prototype: Labcar platform with Lidar, Stereo and Mono camera}
    \label{fig:ugv}
\end{figure}

We developed a prototype of an unmanned ground vehicle (UGV, see, Fig. \ref{fig:ugv}) for conducting our indoor experiment. This platform has a computational module as well as a set of sensors, including 2D lidar, a monocular camera, and ZED stereo camera. The detailed description of the software and hardware used is presented in this section.

\subsection{Hardware and software}
\label{sec:hardware}

The experiment consists of two parts: (1) the dataset acquisition, and (2) the dataset execution from the recorded sensors data. We used Labcar platform based on Traxxas 7407 Radio-Controlled Car Model \cite{shimchik2016,sheikh2018} for the dataset acquisition and a ground station for the dataset execution.  

\begin{figure}[!htbp]
        \captionsetup{font=scriptsize}
		\captionsetup[subfigure]{aboveskip=4pt,belowskip=0pt}
		\begin{subfigure}[b]{0.235\textwidth}
		    \captionsetup{font=scriptsize}
			\centering
		\includegraphics[width=\textwidth]{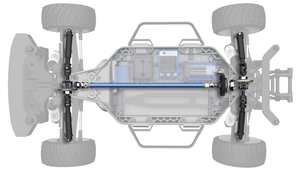} 
			\caption{\textnormal {\fontsize{8}{12}\fontfamily{ptm}\selectfont 4WD Traxxas \#74076}}
			\label{fig:chassis_a}
		\end{subfigure}
		\hfill
		\begin{subfigure}[b]{0.235\textwidth}
		    \captionsetup{font=scriptsize}
			\centering
		\includegraphics[width=\textwidth]{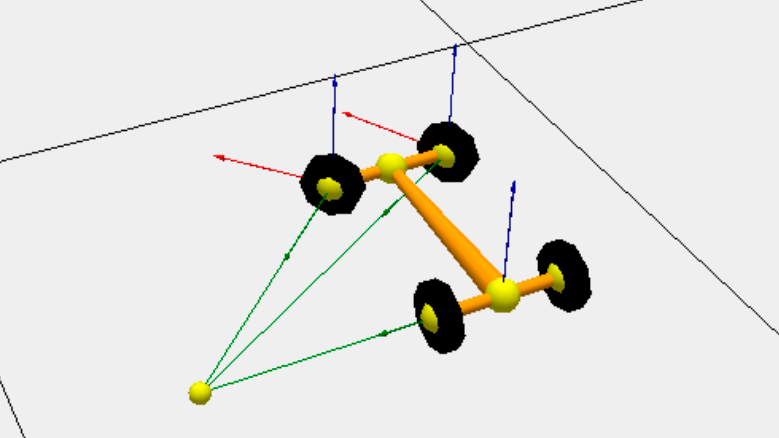} 
			\caption{\textnormal {\fontsize{8}{12}\fontfamily{ptm}\selectfont Ackerman steering model}}
			\label{fig:chassis_b}
		\end{subfigure}
	\caption{\fontsize{8}{12}\fontfamily{ptm}\selectfont Chassis of Labcar UGV based  on  Traxxas Radio-Controlled  Car Model.}
	\label{fig:chassis}
\end{figure}

Labcar platform has chassis with Ackermann steering model, which is presented in Fig. \ref{fig:chassis}. As far as this model does not assume rotations at relatively small speeds, it has a positive impact on robustness of V-SLAM methods (since they could have a bad performance at rotational movements). 
Our Labcar platform uses NVidia Jetson TX1 embedded controller as the computation module with Ubuntu 16.04 operation system and custom ROS Kinetic Kame. The platform includes the following on-board sensors: 2D lidar, a monocular camera, and ZED stereo camera. Detailed information about robot platform is presented in Table \ref{tab:carp}.
~
\begin{table}[!htbp]
    \captionsetup{font=scriptsize}
    \caption{\label{tab:carp}\fontsize{8}{12}\fontfamily{ptm}\selectfont \textsc{HARDWARE AND SOFTWARE SPECIFICATION OF THE LABCAR PLATFORM}}
    \centering
    \begin{tabular}{  p{4cm} | p{4cm} }
    \toprule
    \textbf {PARAMETERS} & \textbf {CONFIGURATION} \\
    \midrule
    Chassi & 4WD Traxxas \#74076 \\
    \hline
    \textbf{Hardware} & Jetson TX1 \\
    \hline
    Processor & Quad ARM® A57 \\
    GPU & NVIDIA Maxwell ™ \\
    RAM & 4 GB \\
    \hline
    \textbf{Sensors} &  \\
    \hline
    Lidar & Hokuyo UTM-30LX \\
    Camera & Basler acA1300-200uc \\ 
    Stereo camera & ZED camera \\ 
    \hline
    \textbf{Software} &  \\
    \hline
    JetPack & 3.1 \\ 
    OS & Ubuntu 16.04 \\ 
    ROS & Kinetic Kame \\ 
    \bottomrule
    \end{tabular}
\end{table}
~
Since we processed our results offline, we used the ground station to run different SLAM system on the dataset data which was acquired from Labcar platform sensors. The detailed information about the ground station is presented at the Table \ref{tab:gstation}.

\begin{table}[!htbp]
    \captionsetup{font=scriptsize}
    \caption{\label{tab:gstation}\fontsize{8}{12}\fontfamily{ptm}\selectfont \textsc{HARDWARE AND SOFTWARE SPECIFICATION OF THE GROUND STATION}}
    \centering
    \begin{tabular}{  p{4cm} | p{4cm} }
    \toprule
    \textbf {Parameters} & \textbf {Configuration} \\
    \midrule
    Processor & Intel® Core™ i7 6500U \\
    GPU & NVIDIA® GeForce® GTX 950M \\
    RAM & 12 GB \\
    \hline
    \textbf{Software} & \textbf{Configuration} \\
    \hline
    OS & Ubuntu 16.04 \\ 
    ROS & Kinetic Kame \\ 
    \bottomrule
    \end{tabular}
\end{table}

\subsection{Robot model}
\label{sec:model}

\begin{figure}[!htbp]
    \captionsetup{font=scriptsize}
		\captionsetup[subfigure]{aboveskip=4pt,belowskip=0pt}
		\begin{subfigure}[b]{0.235\textwidth}
		    \captionsetup{font=scriptsize}
			\centering
		\includegraphics[width=\textwidth]{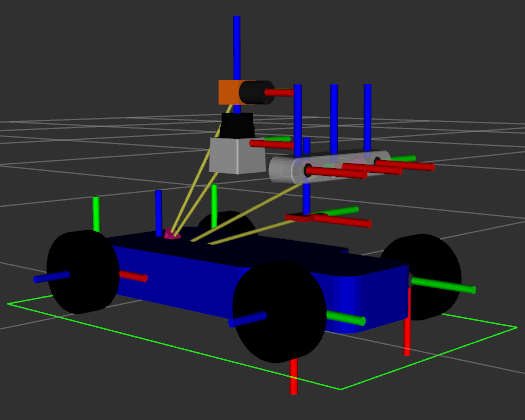} 
			\caption{\textnormal {\fontsize{8}{12}\fontfamily{ptm}\selectfont Robot visualization in RViz}}
			\label{fig:robotmodel_a}
		\end{subfigure}
		\hfill
		\begin{subfigure}[b]{0.235\textwidth}
		    \captionsetup{font=scriptsize}
			\centering
		\includegraphics[width=\textwidth]{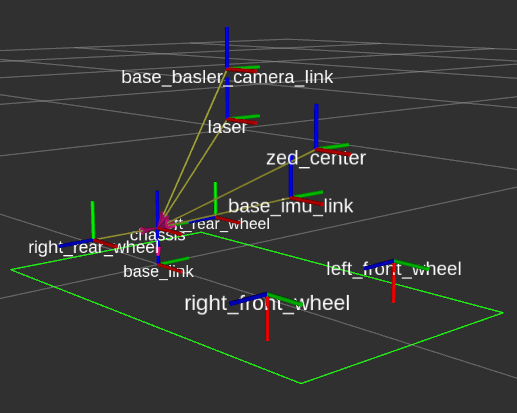} 
			\caption{\textnormal {\fontsize{8}{12}\fontfamily{ptm}\selectfont The model coordinate system}}
			\label{fig:robotmodel_b}
		\end{subfigure}
	\caption{\fontsize{8}{12}\fontfamily{ptm}\selectfont The Labcar platform modelling in ROS/Gazebo simulator environment.}
	\label{fig:robotmodel}
\end{figure}

All tested SLAM systems have integration with ROS (Robot Operating System), that allows to use ROS advantages and improve the experiment quality. Some of them provide full integration, but some of them do this only in form of wrappers. For robot modeling, we applies URDF and \textit{tf} package, transferring all data about UGV positions at the same coordinate frame, and what is more having data from different SLAM systems at the same coordinates with the \textit{base\_link}. It provides also more informative dataset representation and allows reducing human errors at conducting the experiments. The URDF robot model for Labcar platform is presented in Fig. \ref{fig:robotmodel}.
~
~

%% file: Sections/experiment.tex
\section{Methodology of the experiment}
\label{experiment}

\subsection{Experimental environment}
The experiment was conducted with a mobile robot launched within teleoperated closed-loop trajectory along a known perimeter of a rectangular work area for indoor environment of a typical office with monochrome walls. The experimental environment is presented in Fig. \ref{fig:exenv}.

\begin{figure}[!htbp]
    \captionsetup{font=scriptsize}
		\captionsetup[subfigure]{aboveskip=4pt,belowskip=0pt}
		\begin{subfigure}[b]{0.142\textwidth}
		    \captionsetup{font=scriptsize}
			\centering
		\includegraphics[width=\textwidth]{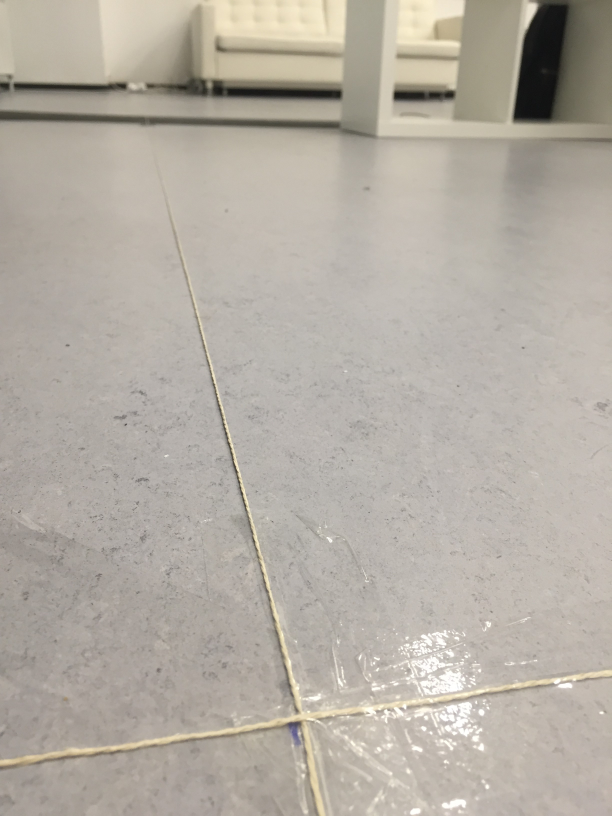} 
			\caption{\textnormal {\fontsize{8}{12}\fontfamily{ptm}\selectfont Marked line}}
			\label{fig:exenv_a}
		\end{subfigure}
		\hfill
		\begin{subfigure}[b]{0.335\textwidth}
		    \captionsetup{font=scriptsize}
			\centering
		\includegraphics[width=\textwidth]{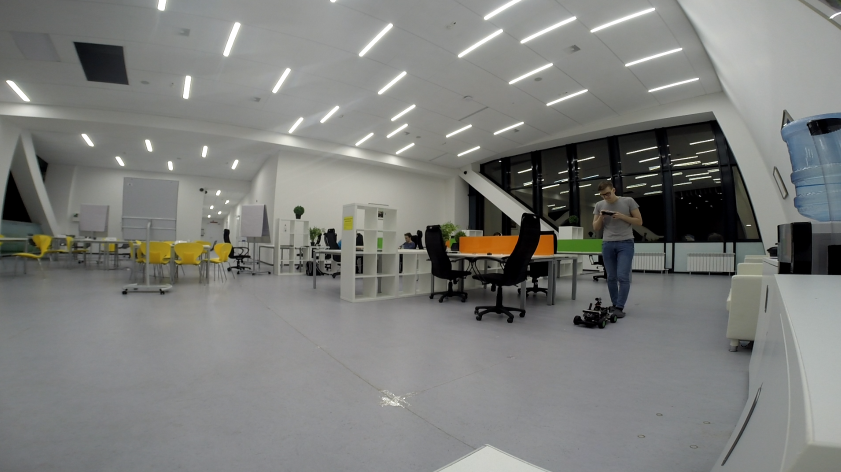} 
			\caption{\textnormal {\fontsize{8}{12}\fontfamily{ptm}\selectfont Test environment}}
			\label{fig:exenv_b}
		\end{subfigure}
	\caption{\fontsize{8}{12}\fontfamily{ptm}\selectfont Indoor environment for the experiment with a mobile robot navigation.}
	\label{fig:exenv}
\end{figure}

\begin{figure}[!htbp]
    \captionsetup{font=scriptsize}
		\captionsetup[subfigure]{aboveskip=4pt,belowskip=0pt}
		\begin{subfigure}[b]{0.18\textwidth}
		    \captionsetup{font=scriptsize}
			\centering
		\includegraphics[width=\textwidth]{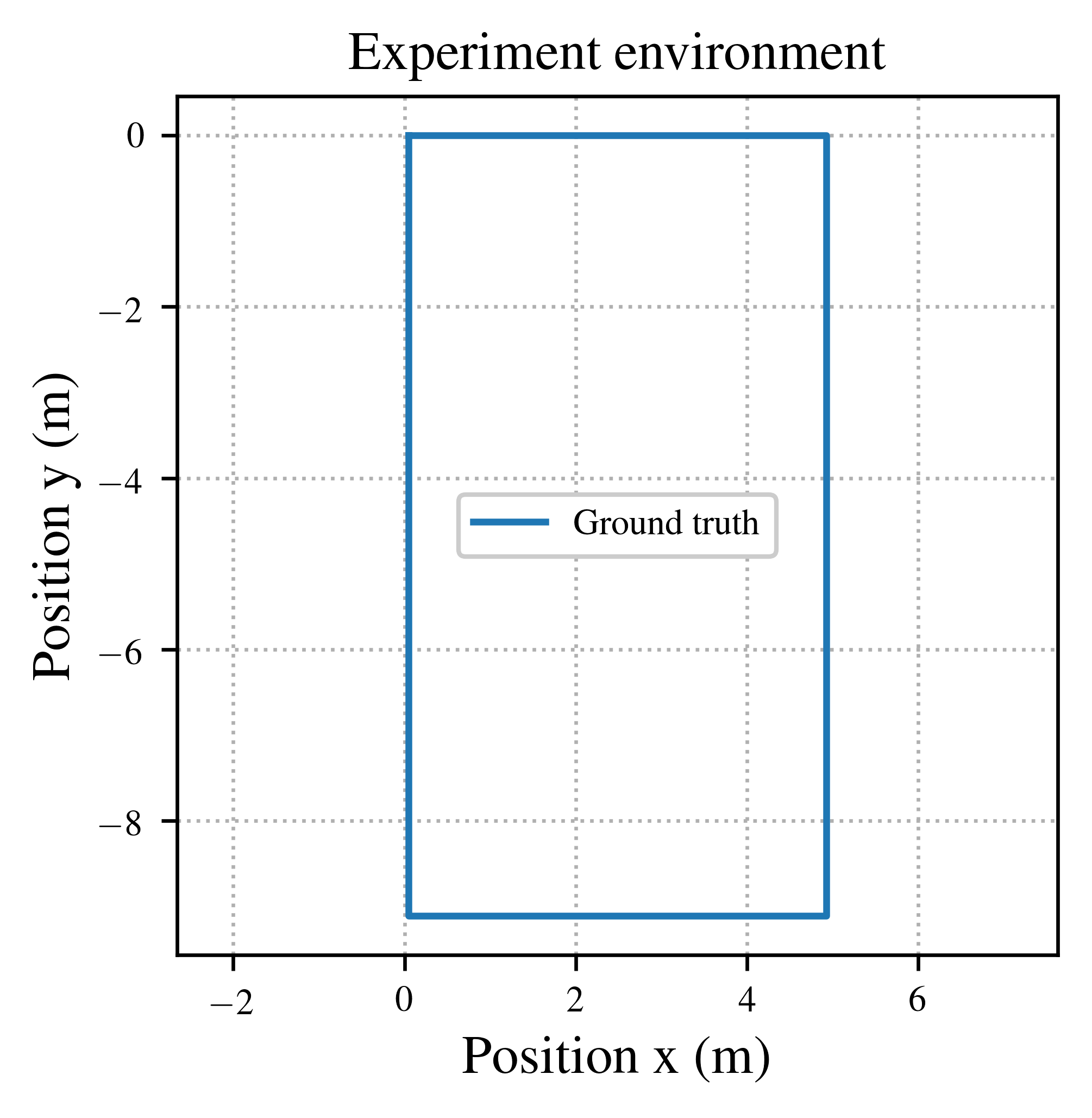} 
			\caption{\textnormal {\fontsize{8}{12}\fontfamily{ptm}\selectfont Test area}}
			\label{fig:line}
		\end{subfigure}
		\hfill
		\begin{subfigure}[b]{0.3\textwidth}
		    \captionsetup{font=scriptsize}
			\centering
		\includegraphics[width=\textwidth]{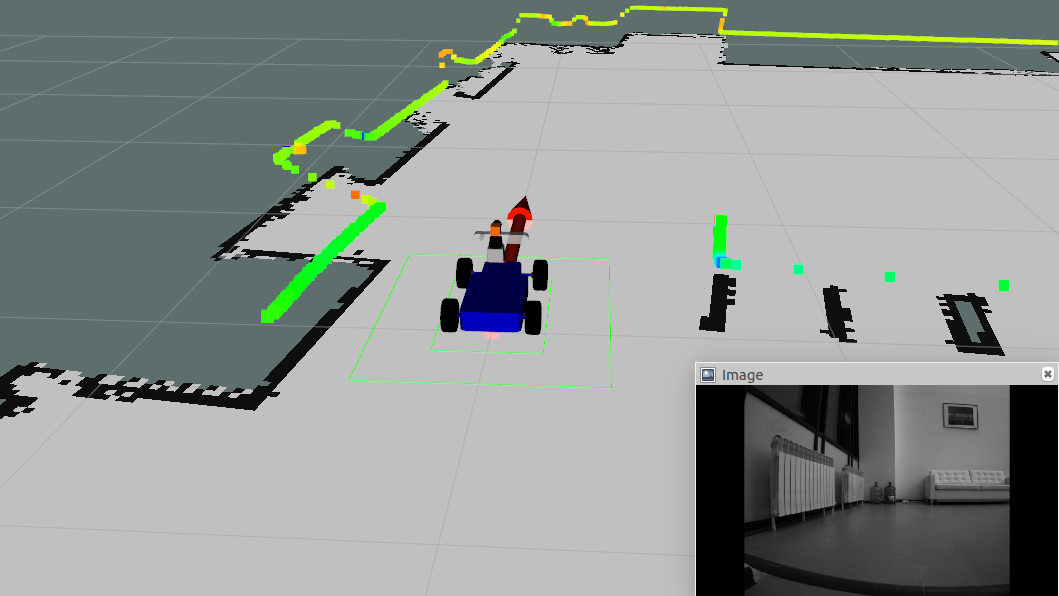} 
			\caption{\textnormal {\fontsize{8}{12}\fontfamily{ptm}\selectfont Lidar \& camera data with robot model}}
			\label{fig:example}
		\end{subfigure}
	\caption{\fontsize{8}{12}\fontfamily{ptm}\selectfont The test polygon model: lidar \& camera data visualization with the robot model in RViz.}
	\label{fig:exenv}
\end{figure}

We used a thread to mark the perimeter of UGV trajectory. The test area is presented in Figure \ref{fig:line}. We refer to this marked line as a Ground truth, although the robot trajectory coincided only during direct movement. The transition between two straight lines occurred with a turn radius of approximately 1 m. Data from sensors was collected in \textit{ROS\_bag} for Basler Camera, ZED stereo camera, and 2D lidar. Ground station was used to run SLAM algorithms based on the collected dataset.

The Fig. \ref{fig:example} shows the visualization of the dataset in RViz, demonstrating the robot model, a window of the forward camera, 2D lidar point cloud and the map built.

\subsection{Metrics for evaluation}
\label{sec:metrics}

We used Absolute Trajectory Error (ATE) for experiment evaluation, comparing the collect data from the tested system with more accurate data about UGV trajectory. We used data from 2D lidar-based Hector SLAM to verify visual SLAM systems, assuming that Hector SLAM provides higher accuracy that will be confirmed further. 

Let $(x_i, y_i, t_i)$ represent $x$ and $y$ coordinates  at time $t_i$ at the Hector SLAM system-based coordinate frame. Then $(x_i^`, y_i^`, t_i^`)$ correspond to data for the exploring method. We interpolate Hector SLAM trajectory with the first degree polynomial. Let denote $x(t)$ and $y(t)$ as interpolated Hector SLAM trajectory data. Then we compute the absolute trajectory error by the following equation \ref{eq:ate}:

\begin{gather} \label{eq:ate}
ATE\mathrm({t}_{i}^{'})=\left\| (x\mathrm({t}_{i}^{'}), y\mathrm({t}_{i}^{'})) - (\mathrm{x}_{i}^{'},\mathrm{y}_{i}^{'}) \right\|
\end{gather}

To represent this evaluation function we use statistical metrics: Root Mean Square Error (RMSE), Mean, Median, Standard deviation (Std.), Minimum, Maximum. Dataset was used to run different ROS-based SLAM system on the ground station, where all metrics were collected and evaluated.

%% file: Sections/evaluation.tex
\section{Evaluation and Discussion}
\label{evaluation}

\subsection{Odometry analysis}
\label{sec:odometry}

The metrics, which evaluate mobile robot localization with different SLAM methods, are presented in Table \ref{tab:allTable}.

\begin{table}[!htbp]
\captionsetup{font=scriptsize}
\caption{\label{tab:allTable}\fontsize{8}{12}\fontfamily{ptm}\selectfont \textsc{ABSOLUTE TRAJECTORY ERROR FOR DIFFERENT SYSTEMS BASED ON HECTOR SLAM TRAJECTORY}}
\centering
\begin{tabular}{ p{1.6cm} | p{0.67cm} | p{0.67cm} | p{0.81cm} | p{0.67cm} | p{0.67cm} | p{0.67cm} }
\toprule
    System & RMSE & Mean & Median & Std. & Min & Max  \\
     & (m) & (m) & (m) & (m) & (m) & (m)  \\
\midrule
    Cartographer & 0.024 & 0.017 & 0.013 & 0.021 & 0.001 & 0.07 \\ 
    LSD SLAM & 0.301 & 0.277 & 0.262 & 0.117 & 0.08 & 0.553 \\ 
    ORB SLAM (mono) & 0.166 & 0.159 & 0.164 & 0.047 & 0.047 & 0.257 \\ 
    DSO & 0.459 & 0.403 & 0.419 & 0.219 & 0.007 & 0.764 \\ 
    ZEDfu & 0.726 & 0.631 & 0.692 & 0.358 & 0.002 & 1.323 \\ 
    RTAB map & 0.163 & 0.138 & 0.110 & 0.085 & 0.004 & 0.349 \\ 
    ORB SLAM (stereo) & 0.190 & 0.151 & 0.102 & 0.115 & 0.004 & 0.414 \\ 
    S-PTAM (no loop cl.) & 0.338 & 0.268 & 0.244 & 0.206 & 0.001 & 0.768 \\ 
    S-PTAM (loop cl.) & 0.295 & 0.257 & 0.242 & 0.145 & 0.006 & 1.119 \\
\bottomrule
\end{tabular}
\end{table}

From the tested SLAM methods, RTAB map can be considered as one of the best method in terms of RMSE for mobile robot localization problem in homogeneous indoor office environment. The monocular ORB SLAM may be estimated as the second most accurate method, whereas ZEDfu demonstrated the worse results for our experiments.

\subsection{Maps analysis}
\label{sec:maps}

\begin{figure}[!htbp]
    \captionsetup{font=scriptsize}
    \centering
    \hfill
    \begin{subfigure}[b]{0.158\textwidth}
        \captionsetup{font=scriptsize}
        \includegraphics[width=\textwidth]{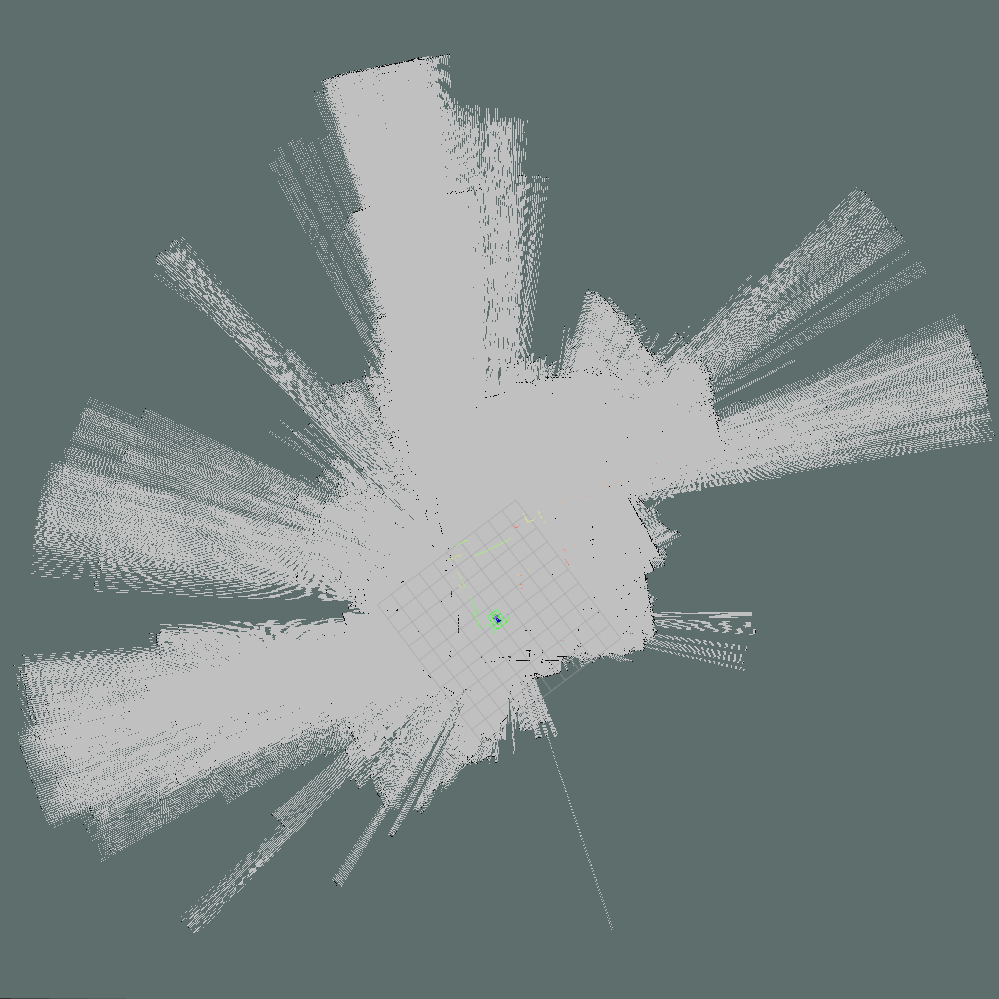}
        \caption{\fontsize{8}{12}\fontfamily{ptm}\selectfont GMapping}
        \vspace{0.5mm}
    \end{subfigure}
    \begin{subfigure}[b]{0.158\textwidth}
        \captionsetup{font=scriptsize}
        \includegraphics[width=\textwidth]{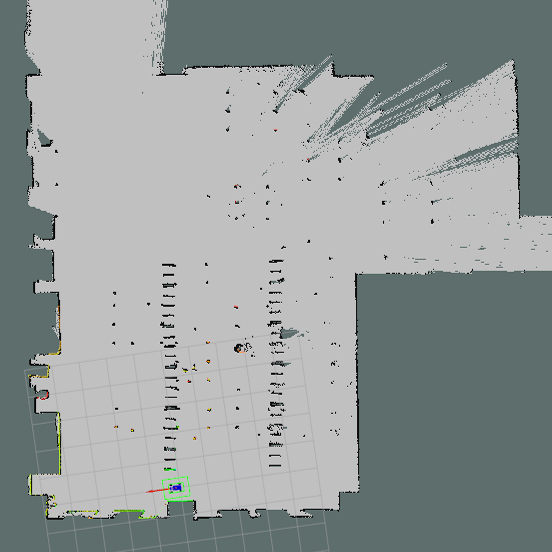}
        \caption{\fontsize{8}{12}\fontfamily{ptm}\selectfont Hector SLAM}
        \vspace{0.5mm}
    \end{subfigure}
    \begin{subfigure}[b]{0.158\textwidth}
        \captionsetup{font=scriptsize}
        \includegraphics[width=\textwidth]{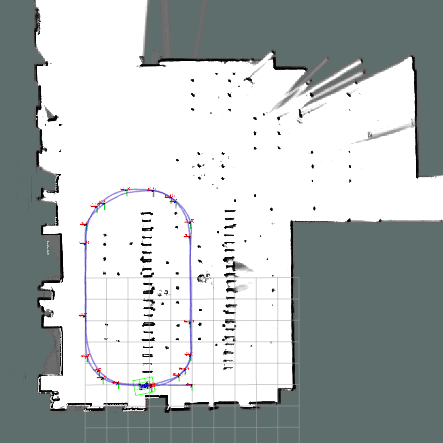}
        \caption{\fontsize{8}{12}\fontfamily{ptm}\selectfont Cartographer}
        \vspace{0.5mm}
    \end{subfigure}
    \hfill
    \begin{subfigure}[b]{0.158\textwidth}
        \captionsetup{font=scriptsize}
        \includegraphics[width=\textwidth]{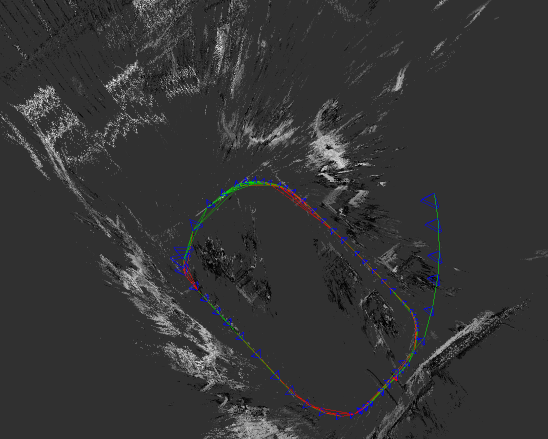}
        \caption{\fontsize{8}{12}\fontfamily{ptm}\selectfont LSD SLAM}
        \vspace{0.5mm}
    \end{subfigure}
    \begin{subfigure}[b]{0.158\textwidth}
        \captionsetup{font=scriptsize}
        \includegraphics[width=\textwidth]{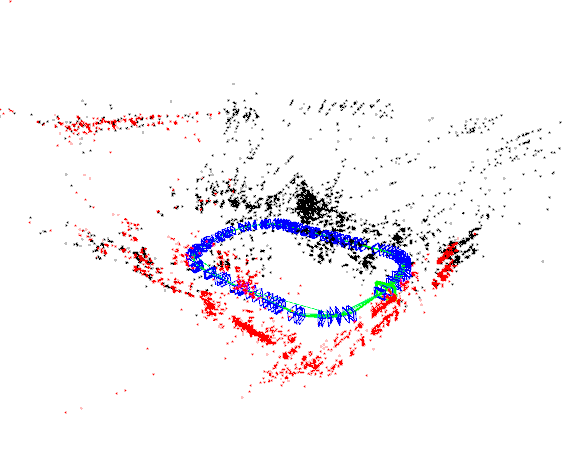}
        \caption{\fontsize{8}{12}\fontfamily{ptm}\selectfont ORB (mono)}
        \vspace{0.5mm}
    \end{subfigure}
    \begin{subfigure}[b]{0.158\textwidth}
        \captionsetup{font=scriptsize}
        \includegraphics[width=\textwidth]{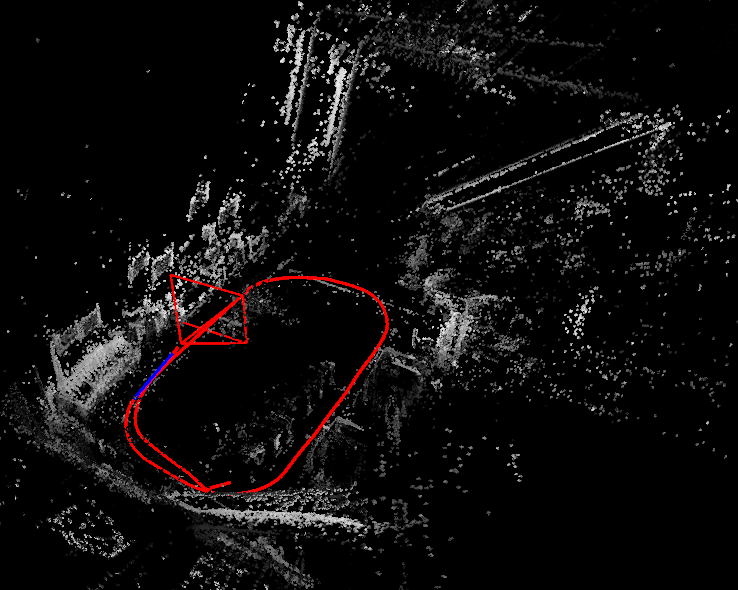}
        \caption{\fontsize{8}{12}\fontfamily{ptm}\selectfont DSO}
        \vspace{0.5mm}
    \end{subfigure}
    \hfill
    \begin{subfigure}[b]{0.158\textwidth}
        \captionsetup{font=scriptsize}
        \includegraphics[width=\textwidth]{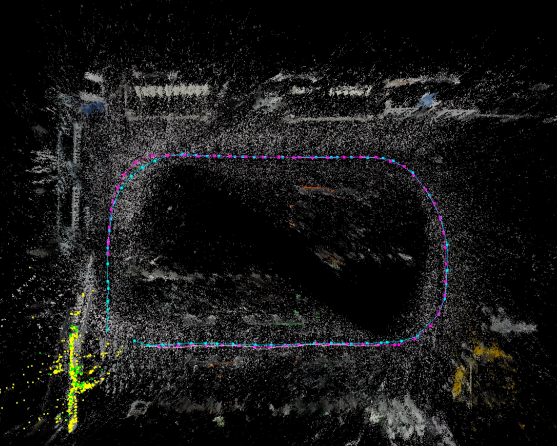}
        \caption{\fontsize{8}{12}\fontfamily{ptm}\selectfont RTAB map}
    \end{subfigure}
    \begin{subfigure}[b]{0.158\textwidth}
        \captionsetup{font=scriptsize}
        \includegraphics[width=\textwidth]{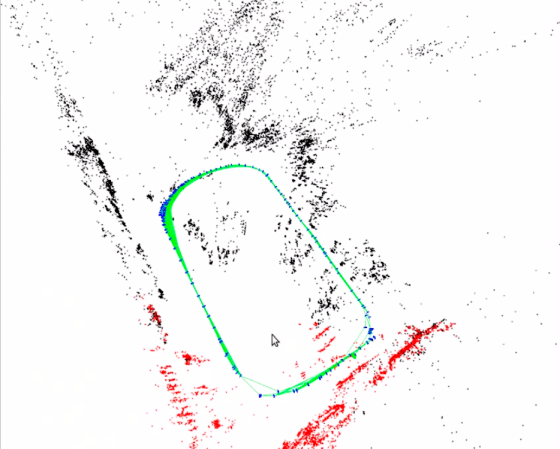}
        \caption{\fontsize{8}{12}\fontfamily{ptm}\selectfont ORB (stereo)}
    \end{subfigure}
    \begin{subfigure}[b]{0.158\textwidth}
        \captionsetup{font=scriptsize}
        \includegraphics[width=\textwidth]{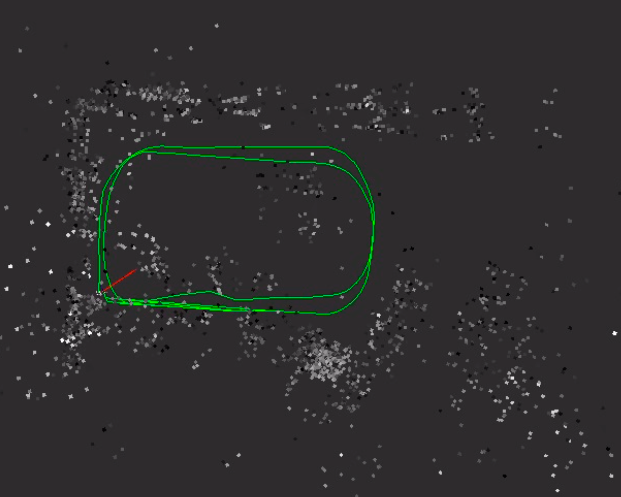}
        \caption{\fontsize{8}{12}\fontfamily{ptm}\selectfont S-PTAM}
    \end{subfigure}
    \caption{\fontsize{8}{12}\fontfamily{ptm}\selectfont Maps generated by various SLAM methods}
    \label{fig:allSLAM}
\end{figure}

The maps generated by different SLAM systems are presented in Figure \ref{fig:allSLAM}.

\textbf{GMapping, Hector SLAM, Cartographer} lidar based methods are provides \textit{nav\_msgs/OccupancyGrid} map. In our experiments, GMapping built an inaccurate  map, whereas Hector SLAM and Cartographer provided quite similar maps. However, it should be noted that since Cartographer uses global map optimization cycle and local probabilistic map updates, it makes this system more robust to environmental changes. Therefore, at present we can recommend Cartographer as the best choice for 2D lidar SLAM. This conclusion is supported by this research \cite{yagfarov2018}.

\textbf{LSD SLAM, DSO} are direct methods, which provide dense point cloud maps and allow to make 3D scene recovery and object detection. 

\textbf{ORB SLAM, RTAB map, S-PTAM} are feature-based methods that build sparse point cloud maps. Therefore, there are difficulties to make 3D scene recovery, but they are good enough for solving navigation problems.

%% file: Sections/conclusion.tex
\section{Conclusion}
\label{conclusion}

As the result of our work the mobile robot prototype to provide experiments in a typical office environment was developed. The robot was launched indoor within teleoperated closed-loop trajectory along a known perimeter of a square work area, recording sensor’s telemetry data for offline processing. Thus, the dataset from on-board sensors (2D lidar, monocular and stereo cameras) was collected and processed using the following SLAM methods: (a) 2D lidar-based: GMapping, Hector SLAM, Cartographer; (b) monocular camera-based: Large Scale Direct monocular SLAM (LSD SLAM), ORB SLAM, Direct Sparse Odometry (DSO); and (c) stereo camera-based: ZEDfu, Real-Time Appearance-Based Mapping (RTAB map), ORB SLAM, Stereo Parallel Tracking and Mapping (S-PTAM). The results of execution for SLAM systems were investigated and compared with the use of metrics.

The main results of our investigations:
\begin{itemize}
\item 2D lidar SLAM systems: Hector SLAM and Cartographer provide accurate solutions for UGV localization and map building. The methods provide almost the same results with RMSE of Absolute Trajectory Error (ATE) at 0.024 m. Both trajectories coincide with the marked line on the floor. However, since Cartographer uses global map optimization cycle and local probabilistic map updates, it makes this system more robust to environmental changes.

\item Monocular visual SLAM systems: Parallel Tracking and Mapping (PTAM), Semi-direct Visual Odometry (SVO), Dense Piecewise Parallel Tracking and Mapping (DPPTAM) failed the experiments since they lost track due to lack of features.

\item  Monocular visual SLAM systems: Large Scale Direct monocular SLAM (LSD SLAM), ORB SLAM, Direct Sparse Odometry (DSO) can be used for solving localization problem with an additional module for scale recovery.

\item  No monocular SLAM system could handle scale ambiguity problem without additional information about environment for scale recovery.

\item  Stereo visual SLAM systems: ZEDfu, Real-Time Appearance-Based Mapping (RTAB map), ORB SLAM, Stereo Parallel Tracking and Mapping (S-PTAM) provide metric information about localization without additional scaling modules, also building 3D metric point cloud. 

\item Visual SLAM system: RTAB map demonstrated the best results for localization problem in our experiments with RMSE ATE of 0.163 m, but it has the problem with the track lost close to monochrome walls. The most robust and stable between tested system is ORB SLAM with RMSE ATE of 0.190 m.

\end{itemize} 

Since the performance of visual SLAM systems strongly depends on computational resources of a mobile robot, our future studies can concern hardware limitations and their influences on efficiency of SLAM applications. 